\documentclass[11pt,a4paper]{article}
\pdfoutput=1 
\usepackage[table]{xcolor}
\usepackage[hyperref]{acl2020}
\usepackage{times}
\usepackage{latexsym}
\usepackage{microtype}
\usepackage{textcomp}
\usepackage{url}
\usepackage{booktabs}
\usepackage{enumitem}
\usepackage{hyphenat}
\usepackage{soul}
\usepackage{mdframed}
\usepackage{subcaption}
\usepackage{multicol}
\usepackage{tabularx}
\usepackage{float}
\usepackage{array}
\usepackage{makecell}
\usepackage{tikz}
\usetikzlibrary{arrows,positioning,matrix} 
\usepackage{bigstrut}

\aclfinalcopy 
\def\aclpaperid{556} 

\newcommand{\citetb}[1]{\setcitestyle{square}\unskip\citet{#1}\setcitestyle{round}\unskip}

\newcommand{\egtext}[1]{\textit{#1}}
\newcommand{\newterm}[1]{\textsc{#1}}

\DeclareTextFontCommand{\emph}{\bf}

\newcommand{\rightanswer}[1]{{\itshape #1}}

\makeatletter
\newcommand*{\@rowstyle}{}
\newcommand*{\rowstyle}[1]{
  \gdef\@rowstyle{#1}%
  \@rowstyle\ignorespaces%
}
\newcolumntype{=}{
  >{\gdef\@rowstyle{}}%
}
\newcolumntype{+}{
  >{\@rowstyle}%
}
\makeatother

\newcommand{\nameandcite}[1]{#1 \citep{#1}}
\title{To Test Machine Comprehension, Start by Defining Comprehension}

\renewcommand*{\thefootnote}{*}
\author{\parbox{\linewidth}{\centering
            Jesse Dunietz\footnotemark[1], Gregory Burnham\footnotemark[1], Akash Bharadwaj, Owen Rambow, \\
            Jennifer Chu-Carroll, \and David Ferrucci} \\
        Elemental Cognition \\
        \texttt{\{jessed,gregb,akashb,owenr,jenniferc,davef\}}\\\texttt{@elementalcognition.com} \\}

\date{}

\begin{document}
\maketitle
\footnotetext{\label{fn:contribs}Equal contributions.}
\renewcommand*{\thefootnote}{\arabic{footnote}}

\begin{abstract}
Many tasks aim to measure \newterm{machine reading comprehension} (MRC), often focusing on question types presumed to be difficult. Rarely, however, do task designers start by considering what systems should in fact comprehend. In this paper we make two key contributions. First, we argue that existing approaches do not adequately define comprehension; they are too unsystematic about what content is tested. Second, we present a detailed definition of comprehension\textemdash a \newterm{Template of Understanding}\textemdash for a widely useful class of texts, namely short narratives. We then conduct an experiment that strongly suggests existing systems are not up to the task of narrative understanding as we define it.
\end{abstract}

\section{Introduction}
Over the past few years, neural models \citep[e.g.,][]{CNNDM-done,BERT,RoBERTa} have begun to match or even exceed human performance on \newterm{machine reading comprehension} (MRC) benchmarks. In these tasks, systems demonstrate their comprehension of a passage by answering questions about it. Yet despite recent successes, MRC appears far from solved: systems continue to make basic, sometimes baffling mistakes, and they fail to generalize to new data. Such shortcomings have motivated a flurry of new MRC tasks, each designed to confront systems with questions deemed challenging for current methods. For example, tasks may ask questions requiring commonsense reasoning \citep{CosmosQA}, multi-hop reasoning \citep{WikiHop}, or inferences based on a second passage \citep{ROPES}.

This line of research assumes that ever-more-``difficult'' question-answering tasks will ultimately lead to more robust and useful reading comprehension. We argue that, while the question-answering format can be a fine choice for \emph{how} to test comprehension, using difficulty as the basis for \emph{what} to test is fundamentally flawed. To put it provocatively, the dominant MRC research paradigm is like trying to become a professional sprinter by glancing around the gym and adopting any exercises that look hard. The training may end up exercising some relevant muscles, but it is far too haphazard to achieve the ultimate goal.

Like athletic training, MRC tasks are not an end in themselves; ultimately, they are meant to lead to real-world applications. Current tasks may suffice for sufficiently similar applications\textemdash e.g., chatbots that look up customer questions in product documentation. But many proposed NLP applications hinge on deeper comprehension. Early work \citep[e.g.,][]{MGDyerThesis} pointed to examples like assistance with legal disputes and service contracts; more recent work suggests applications such as summarizing a patient's clinical timeline \citep{AllenDeepNLUHealthcare}. For such complex applications, machines will need to manipulate rich models of the world evoked by the text\textemdash e.g., to compare a claimant's narrative to legal standards, or to build a causal model of a patient's condition. From this broader perspective, the current paradigm falls short.

Specifically, we claim that in the quest for difficulty, task designers overlook the issue of what \emph{content}\textemdash what information expressed, implied, or relied on by the passage\textemdash systems should comprehend. MRC datasets are usually constructed by having humans cast about for supposedly tricky questions, most often questions based on reasoning. But the questions that result are scattershot, offering little assurance that even a high-scoring system has achieved a useful and robust understanding. 

We advocate for a different approach. We propose that the first step in defining MRC tasks should be specifying what content a system would likely need to understand for a given class of applications. Only then can tasks systematically compile questions to probe for the internal model that the machine ought to have constructed.

This paper demonstrates such an approach for applications that involve understanding narratives.\footnote{We will use ``narrative'' and ``story'' interchangeably, roughly following the Wikipedia definition: ``A narrative or story is an account of a series of related events, experiences, or the like, whether true\ldots or fictitious.''} After reviewing existing approaches to constructing MRC datasets (\S\ref{sec:lit-review}), we argue for narratives as a valuable MRC testbed (\S\ref{sec:why-stories}). Then, inspired by cognitive science research on reading comprehension, we propose a ``template of understanding'' (ToU) for stories\textemdash an account of what an internal model of a story should minimally contain (\S\ref{sec:tou}). We also suggest ways to operationalize our ToU as a story comprehension task (\S\ref{sec:operationalizing}). Finally, we show evidence from a pilot ToU-based task that current MRC models are not up to the challenge (\S\ref{sec:experiments}).

\section{Existing MRC dataset designs} \label{sec:lit-review}
This paper addresses how MRC tests can be made more systematic. Accordingly, we review existing tasks grouped by their data collection methods. We argue that each category falls short of testing a useful body of content in a satisfying way.

\subsection{Manually written questions}
\label{sec:manual}
By far the most popular strategy for generating MRC questions is to have humans\textemdash usually crowd workers, but sometimes trained annotators\textemdash think of questions about each passage.

The most straightforward version of this method gives annotators little to no guidance regarding what questions to ask. One early example is the TREC-8 dataset \citep{TREC}. 
In the more recent SNLI \citep{SNLI} and MNLI \citep{MNLI} entailment tasks, the only constraint on crowd workers was that they produce one entailed, one contradicted, and one neutral hypothesis for each premise sentence.\footnote{Parts of the original RTE datasets \citep[etc.]{RTE} were generated more systematically, but only in the sense that the outputs of NLP tools (e.g., translation or information extraction systems) were recorded as correct/incorrect examples of entailment. Little attention was paid to subject matter.} Similarly, the workers who assembled \nameandcite{NewsQA} were told only that the questions had to be answerable with short phrases, and workers for \nameandcite{SQuAD} were simply given a ``good'' and a ``bad'' example and encouraged to use original wording.

The problem with such an open-ended generation process is that, absent stronger guidance, people tend to write simple questions that can be answered using lexical cues. (See, e.g., the dataset analysis in \citealp{SQuAD}.) This makes the tasks questionable measures of comprehension.

The dominant solution is to incorporate trickier twists. \nameandcite{NarrativeQA} and \nameandcite{DuoRC} reduce lexical similarity between questions and passages by showing annotators only a second passage about the same events. Other datasets emphasize reasoning presumed to be difficult, such as incorporating information from multiple parts of the text. \nameandcite{MCTest} and \nameandcite{MultiRC} ask for questions that rely on multiple sentences; \nameandcite{ROPES} has annotators apply information from one passage to write questions on a second; and \nameandcite{HotpotQA} and \nameandcite{QASC} require multi-hop reasoning. Other forms of reasoning tested include coreference resolution (Quoref, \citealp{Quoref}; Winograd Schema Challange, \citealp{WSC}), numerical reasoning \citep[DROP,][]{DROP}, and commonsense reasoning \citep[Cosmos QA,][]{CosmosQA}. Tasks can also be made harder with devices such as unanswerable questions (SQuADRUn, \citealp{SQuAD-2}; NewsQA; CosmosQA) and filtering questions with an adversarial baseline (DROP; Quoref; QASC).

These twists do make MRC harder. But to pursue hard questions is to overlook why easy questions seemed inadequate in the first place: MRC tasks are a means to an end, namely useful applications, and easy questions\textemdash e.g., questions that depend only on lexical cues\textemdash do not suffice for that end. 
The techniques above may help by guiding annotators to a different space of questions: intuition suggests that some of these harder questions are indeed useful ones. But such techniques are an incomplete solution, as difficulty is a weak proxy for utility. 
What matters is not the system's sophistication per se; it is the alignment between the questions the system can answer and the ones a given application needs it to. Designing for difficulty still gives little assurance of such alignment.

Perhaps a truly random walk through question space would eventually cover a representative set of useful questions, but annotators are biased toward questions that humans find interesting \citep[see][]{ReportingBiasKA, ReportingBiasVisual, CommonSenseInference}. They do not think to ask questions whose answers seem obvious, even when those answers are essential to comprehension. If we do not delineate such facts and evaluate systems' ability to manipulate them, we will never be satisfied that the systems have adequately understood the text.

\subsection{Naturally occurring questions} \label{sec:naturally-occurring}
A second approach is to find questions ``in the wild,'' then retrospectively collect documents containing the answers. This is the approach of \nameandcite{BoolQ} and MS MARCO \citep{MARCO}, which compile search engine queries, and of \nameandcite{ELI5}, which harvests questions from Reddit's ``Explain Like I'm Five'' forum.

Such datasets are clearly useful for answering common queries, a valuable application class in its own right. For more complex applications, however, common queries are, if anything, less thorough than annotators at probing important elements of understanding (particularly aspects humans find obvious). The mismatch between questions and passage content is exacerbated by finding the passages retrospectively: the questions do not even attempt to test most of what each passage discusses, making them an insufficient measure of MRC.

\subsection{Questions from tests designed for humans}
The third strategy is to pull questions from tests written for humans. Examples include the early ``Deep Read'' corpus \citep{DeepRead}; the more recent \nameandcite{TriviaQA} and \nameandcite{SearchQA} datasets, which mine collections of trivia questions; the AI2 Reasoning Challenge \citep[ARC;][]{ARC}, which asks questions from standardized science tests; and \nameandcite{RACE}, which draws from English learning materials for Chinese school students.

Our chief concern about this approach echoes our concerns from \S\ref{sec:manual}: tests designed for humans rarely bother to test content that most humans find obvious. Accordingly, they gloss over vast swaths of understanding that machines do not yet have but which may be critical to applications. In addition, 
 SearchQA, TriviaQA, and ARC find passages retrospectively, so again, the questions they ask only tangentially graze the content of each passage.

\subsection{Automatically generated questions}
Several projects generate questions algorithmically. The \textit{CNN}/\textit{Daily Mail} datasets \citep{CNNDM} and \nameandcite{ReCoRD} produce cloze-style questions over news passages by masking out entities from summaries and below\hyp{}the\hyp{}fold sentences. ComplexWebQuestions \citep[CWQ;][]{ComplexWebQuestions} and \nameandcite{WikiHop} test for multi-hop reasoning by walking a structured knowledge base. Finally, \nameandcite{bAbI} generates short texts and questions from a simple simulation of characters moving around.

Each algorithm encodes assumptions about what is worth asking. In theory, then, the algorithmic approach could produce a satisfying MRC test: given appropriate inputs, the algorithm could aim to generate questions that cover important content. Indeed, our proposal in \S\ref{sec:annotators} can be seen as a question generation algorithm to be run by humans.

In practice, however, algorithmic approaches have de-emphasized content. \textit{CNN}/\textit{Daily Mail} and ReCoRD capture explicit assertions about maskable entities, which do not amount to a principled body of content. The algorithms behind CWQ and WikiHop at least take as input some body of content, namely knowledge graphs. But the graphs include only a fraction\textemdash again, not a principled one\textemdash of the associated documents' content, and the questions are further restricted to rely on multi-hop reasoning. Multi-hop reasoning is no doubt a major error source for MRC, but applications are driven by what propositions must be extracted; whether each proposition takes zero inference steps or seven is immaterial. Accordingly, multi-hop questions are worth investigating, but they are not a sufficiently well-motivated body of content to constitute a measure of reading comprehension.

Similar remarks can be made about most of bAbI's 20 ``tasks'': grounded in simulations, their question generation algorithms start from known content, but target forms of reasoning. However, the tasks concerning time, positions, sizes, pathfinding, and motivations are closer to our content-first question generation strategy. These tasks are not driven by applications, and their synthetic passages are unrealistically simple, but among existing datasets, they are closest 
to our proposal.

\subsection{Summary: What is missing}
The most clear-cut way to test reading comprehension would be to select passages, describe what should be comprehended from them, and design tests for that understanding. Yet few MRC datasets have even approximated this approach. Many impose little structure on what content is tested; the rest pick some ``difficult'' form(s) of analysis or linguistic phenomena, but rarely consider downstream goals to determine what the questions should be \emph{about}. Metrics for difficult reasoning and linguistic phenomena \citep[see, e.g.,][]{MoreComprehensive} are useful, but only as tools for error analysis and mitigation; they are not top-line performance metrics.

In addition, many datasets to date suffer from two other problems: 1) they select passages after the questions are asked, meaning the questions test comprehension of only small portions of the passages; and/or 2) they ask very few questions whose answers are obvious to humans.


These issues of content scope also intersect with issues of format. Many tasks have adopted a span extraction format, including TREC QA, NewsQA, and (most notably) SQuAD and its successors. This format immediately rules out questions about inferred events or entities, which may be essential to a complete interpretation.
The main alternative is multiple choice (MC), used in tasks such as Cosmos QA, RACE, ARC, WikiHop, and every task in GLUE \citep{GLUE} and SuperGLUE \citep{SuperGLUE}. But MC has its own problem of providing extra hints via answer choices.

We will return to the format issue in \S\ref{sec:operationalizing}. But first, we propose a more systematic approach to constructing MRC datasets.

\section{Defining deep story understanding}
Our approach starts from the \emph{content} of a passage, which we define as the information it expresses, implies, or relies on. Specifically, we propose that task designers lay out a minimal body of content that MRC systems should demonstrate they understand. Exactly what that content is will vary from passage to passage, of course, but the key is to define a \newterm{template of understanding} (ToU): a set of question templates that can be filled in with specific events and entities for any given passage. The answers to the fleshed-out questions will constitute a floor of understanding for the passage\textemdash a plausible lower bound on what content machines ought to comprehend.

The natural next question is what content the ToU should cover. System needs will vary by application. To advance MRC writ large without limiting ourselves to a single application, we propose selecting a class of texts where one could reasonably predict a priori what content would be useful for applications. In the rest of this section, we endorse fictional narratives as a particularly promising class of texts and propose a ToU for them.\footnote{To be clear, we are \emph{not} claiming that fictional narratives are themselves an application; only that they are a class of texts that are useful for many applications.}

\subsection{The case for stories} \label{sec:why-stories}
Stories 
have several convenient properties that recommend them as a testbed for MRC.

Most importantly, applications that involve comprehending stories are numerous and diverse. Consider a legal aid tool: to assess whether a lawsuit may be warranted, it would have to comprehend an account of the events in question. Likewise, a tool that finds candidates for medical trials would need to read each patient history. (Appendix \ref{sec:tou-examples} fleshes out these scenarios.) These examples are not exceptional; applications in other domains will depend on stories in customer complaints, intelligence dispatches, financial news, and many other document types. Humans tend to think and communicate in terms of stories \citep[see, e.g.,][]{RighteousMindQuote, NarrativeIntelligence, NarrativeCxn, StorytellingInPedagogy}, so it is unsurprising that stories are ubiquitous in the content we want NLU tools to help us with.

Additionally, stories come with a strong prior from cognitive science about what elements of understanding will be useful. Research on human reading comprehension \citep[e.g.,][]{HumanInferences,RCSituationModels} suggests that humans attend primarily to the timeline of events, to the locations of entities and events, and to the causes and motivations of events and actions. For applications that involve story comprehension, we can expect that machines will need to understand these same dimensions.
We can thus design a principled ToU for stories even without specifying an application.

Stories' content also makes them a particularly compelling demonstration of understanding, for two reasons. First, cognitive science suggests that humans make more inferences when reading narrative text than expository text \citep{HumanInferences}. In particular, a story entails a highly structured network of relations (timelines, causality, etc.). 
Thus, stories do exercise abilities beyond simple factoid extraction.
Second, stories rely  on a large body of implicit world knowledge. If a system is able to use and express that knowledge when reading stories, it will likely be able to apply the same knowledge even when comprehending other kinds of texts.

Among stories, fictional ones offer the strongest test of comprehension: their contents cannot be found in corpora, so systems must rely on comprehending the text \citep{MCTest}. Accordingly, we suggest using fictional narratives as the basis for developing a ToU and evaluating MRC.

\subsection{A ToU for stories} \label{sec:tou}
We propose four overlapping clusters of questions for story comprehension, corresponding to the four elements identified by \citet{RCSituationModels} as the ones humans attend to when reading stories.  Further support for these questions, particularly the last two, comes from early work in computational story understanding: \citet{SchankAbelson} identify causal chains, plans and goals as crucial elements of understanding multi-sentence stories.
\begin{enumerate}[itemsep=0pt]
    \item \textbf{Spatial:} Where are entities positioned over time, relative to landmarks and each other? How are they physically oriented? And where do events take place?
    \item \textbf{Temporal:} What events and sub-events occur, and in what order? Also, for what blocks of that timeline do entities' states hold true?
    \item \textbf{Causal:} How do events and states lead mechanistically to the events and states described or implied by the text?
    \item \textbf{Motivational:} How do agents' beliefs, desires, and emotions lead to their actions?
\end{enumerate}
These question templates form the ToU. Systems should ideally be able to answer them about all entities and events that the story mentions or implies (though of course some entities/events are more important than others; see \S\ref{sec:annotators}). We do not have a separate category for ``who did what to whom'' information, but we expect strong performance on the ToU to hinge on such analysis. In particular, much of this information is captured in the characterization of events for temporal questions.

Of course, these four facets do not cover everything one might comprehend. They include nothing about the story's message, or how it resembles other stories, or even most counting questions. The ToU merely provides a lower bound on what is needed. That said, many forms of reasoning (e.g., counting) can be reduced to deterministically manipulating the answers to multiple ToU questions.

\section{Towards a story understanding task} \label{sec:operationalizing}
Our ToU provides a conceptual framework for stating what a machine should understand from a story. However, there remains the challenge of operationalizing the framework\textemdash i.e., of rigorously assessing whether a machine has that understanding.

We do not claim to have solved this problem, but in this section we discuss two broad directions for further development: evaluating based on annotated answers to ToU questions and asking untrained humans to rank different answers. These approaches might even be combined to offer complementary perspectives on system  performance.

\subsection{Approach 1: Annotating ToU answers} \label{sec:annotators}
One class of approaches starts with trained annotators writing plain-English answers to each ToU question. The annotators are given guidelines for instantiating the ToU on new stories and for making answers detailed and thorough. We call an annotator's answer document a \newterm{record of understanding} (RoU); see Figure \ref{fig:RoU-example} for an example.

\begin{figure}[!tbp]
    \setlist[itemize]{nosep,itemsep=2pt,topsep=2pt,leftmargin=17pt}
    \small
    \begin{mdframed}
        %
        \textbf{Spatial} \textit{(sample entries)}:
        \begin{itemize}
            \item Rover is in the yard from when he runs out the door until he runs inside.
            \item Rover is in the house from when he runs inside until the end of the story.
        \end{itemize}

        \vspace{0.35em}\textbf{Temporal} \textit{(sample entries)}:
        \begin{itemize}
            \item Allie arrives just before Rover runs outside.
            \item Rover barks just before he runs inside.
            \item It is still raining at the end of the story.
        \end{itemize}

        
        \vspace{0.35em}\textbf{Motivational} \textit{(sample entry)}:
        \begin{itemize}
            \item Rover runs inside, rather than staying put, because:
            \begin{itemize}[leftmargin=*]
                \item If he runs inside, he will be inside, whereas if he does not he will be outside, because:
                \begin{itemize}[leftmargin=*]
                    \item Rover is outside.
                    \item Running to a place results in being there.
                \end{itemize}
                \item If Rover is inside, he will not get rained on, whereas if he is outside he will, because:
                    \begin{itemize}[leftmargin=*]
                        \item It is raining.
                        \item When it is raining, things that are outside tend to get rained on, whereas things inside do not.
                    \end{itemize}
                \item Rover would prefer not getting rained on to getting rained on, because:
                    \begin{itemize}[leftmargin=*]
                        \item Most dogs prefer not to get rained on.
                    \end{itemize}
            \end{itemize}
        \end{itemize}
    \end{mdframed}
    \caption{A partial RoU for the following simple story fragment: \egtext{\ldots One day, it was raining. When Allie arrived, Rover ran out the door. He barked when he felt the rain. He ran right back inside.}}
    \label{fig:RoU-example}
\end{figure}

Conceptually, answering temporal and spatial questions is straightforward, but the causal and motivational questions require more definition. People accept many kinds of answers to such questions. It is therefore important to clarify what a good answer should include\textemdash i.e., what causal or motivational facts an MRC system should comprehend.

We base our account of these questions on the philosophical literature on causality \citep[see][]{SEP-causation} and on the social science literature on what explanations people seek \citep[see][]{AI-explicability-SS}. Following this scholarship, we conceptualize a causal or motivational question as asking what root cause led the event or state from the story to happen rather than some alternative outcome. For example, in a story about Rover the dog, the question of why Rover came inside is taken to mean: \egtext{Why did Rover come inside, rather than remaining where he was?}\footnote{Causality as contrast may seem unintuitive, particularly since ``why'' questions tend not to state a contrasting outcome. But the audience generally just infers a reasonable default.

Beyond its support in the literature, contrast offers several advantages. It makes it far easier to match intuitions about what should factor into a causal explanation. It also naturally handles relative preferences, and allows explaining multiple aspects of an event\textemdash e.g., John walking carefully can be explained in contrast to both staying put and walking normally.}

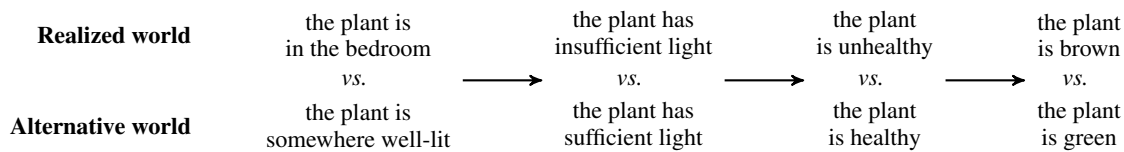
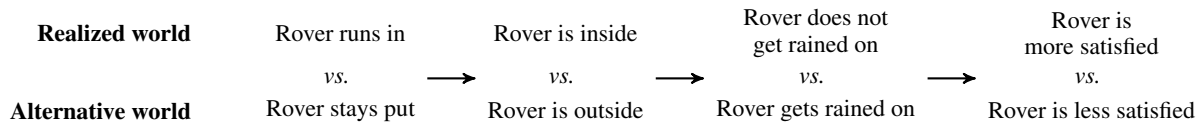
\begin{figure*}[!tpb]
    \centering
    \small
    \tikzset{
        >=stealth',
        every path/.style={
            ->,
            draw,
            thick,
            shorten <=2pt,
            shorten >=2pt
        }
    }
    
    \begin{subfigure}[b]{\textwidth}
        \begin{tikzpicture}[align=center]
            \matrix (m) [matrix of nodes, nodes in empty cells, row sep=0pt, column sep=0em,
                    column 1/.style={anchor=east},
                    column 2/.style={anchor=center},
                    column 3/.style={anchor=center, text width=3em}, 
                    column 4/.style={anchor=center},
                    column 5/.style={anchor=center, text width=3em},
                    column 6/.style={anchor=center},
                    column 7/.style={anchor=center, text width=3em},
                    column 8/.style={anchor=center},
            ] {
                \node {\textbf{Realized world} \hspace{2.2em}};
                    & \node {the plant is\\in the bedroom}; &
                    & \node {the plant has\\insufficient light}; &
                    & \node {the plant\\is unhealthy}; &
                    & \node {the plant\\is brown}; \\
                & \textit{vs.} & & \textit{vs.} & & \textit{vs.} & & \textit{vs.} \\
                \node {\textbf{Alternative world} \hspace{2.2em}};
                    & \node {the plant is\\somewhere well-lit}; &
                    & \node {the plant has\\sufficient light}; &
                    & \node {the plant\\is healthy}; &
                    & \node {the plant\\is green}; \\
            };
            
            \draw (m-2-3.west) -- (m-2-3.east);
            \draw (m-2-5.west) -- (m-2-5.east);
            \draw (m-2-7.west) -- (m-2-7.east);
        \end{tikzpicture}
        \caption{A mechanistic causal chain for the question, ``Why did the plant turn brown?''}
        \label{fig:Zoey}
    \end{subfigure}

    \hrulefill

    \begin{subfigure}[t]{\textwidth}
        \centering
        \begin{tikzpicture}[align=center]
            \matrix (m) [matrix of nodes, nodes in empty cells, row sep=0pt, column sep=0em,
                    column 1/.style={anchor=east},
                    column 2/.style={anchor=center},
                    column 3/.style={anchor=center, text width=1.7em}, 
                    column 4/.style={anchor=center},
                    column 5/.style={anchor=center, text width=1.7em},
                    column 6/.style={anchor=center},
                    column 7/.style={anchor=center, text width=1.7em},
                    column 8/.style={anchor=center},
            ] {
                \node {\textbf{Realized world} \hspace{2.2em}};
                    & \node {Rover runs in}; &
                    & \node {Rover is inside}; &
                    & \node {Rover does not\\get rained on}; &
                    & \node {Rover is\\more satisfied}; \\
                & \textit{vs.} & & \textit{vs.} & & \textit{vs.} & & \textit{vs.} \\
                \node {\textbf{Alternative world} \hspace{2.2em}};
                    & \node {Rover stays put}; &
                    & \node {Rover is outside}; &
                    & \node {Rover gets rained on}; &
                    & \node {Rover is less satisfied}; \\
            };
            
            \draw (m-2-3.west) -- (m-2-3.east);
            \draw (m-2-5.west) -- (m-2-5.east);
            \draw (m-2-7.west) -- (m-2-7.east);
        \end{tikzpicture}

        \caption{A motivational causal chain for the question, ``Why did Rover the dog run back inside when it started raining?''}
        \label{fig:Rover}
    \end{subfigure}
    
    \caption{Example causal chains answering causal (above) and motivational (below) ToU questions.}
    \label{fig:causal-chains}
\end{figure*}

The answer to such a question is a \newterm{causal chain} tracing from the root cause to the event or state described in the story (see Figure \ref{fig:causal-chains} for examples). The links in the chain walk in lockstep through two parallel worlds: the \newterm{realized world}, where the root cause held true and led to the observed outcome; and an \newterm{alternative world}, where the root cause would have been changed and led to some alternative outcome.

For mechanistic causation, each link in the chain ends in an event that helped bring about the outcome described in the story. For example, two mechanistic links from Figure \ref{fig:Zoey} are \egtext{the plant looks brown (rather than green) because it is unhealthy (rather than healthy)} and \egtext{the plant is unhealthy because it has little light (rather than lots of light)}.

For motivations, the structure is slightly different. Rather than the final link being an event that happened in the story, it is a statement of the agent's preferences (in Figure \ref{fig:Rover}, \egtext{Rover would prefer not being rained on to being rained on}). The links leading to it are the future causes and effects that the agent imagines will lead from their action to their preferred outcome (e.g., going inside leading to being inside leading to not getting rained on).

The causal chain provides the backbone of an explanation for an event or action, but the full explanation should recursively explain each link (e.g., \egtext{Rover would prefer not being rained on to being rained on}). Recursive explanations appeal to some combination of general knowledge about the world (e.g., \egtext{Most dogs prefer not to get rained on}) and story-specific \newterm{supporting facts}\textemdash e.g., the fact that Rover is outside. Supporting facts generally need to be recursively explained, as well.


Even with guidelines, different annotators may give substantively different answers. In particular, they may drill down to different levels of detail in a causal chain before bottoming out in general knowledge\textemdash e.g., rather than stopping at dogs disliking rain, one annotator might explain that Rover disprefers rain because he dislikes getting wet, which in turn is because dogs often dislike getting wet. To handle such disagreements, we can adopt the pyramid method \citep{Pyramid} from abstractive summarization, another task where annotators may provide different but equally sensible ground truths. Under this method, a reconciler merges RoUs into a single rubric by identifying shared content ``nuggets'' (e.g., that it is raining) and weighting each by how many annotators cited it. (See \citetb{nuggets} for more on nuggets.)

\subsubsection{Preliminary notes on RoU agreement}
\label{sec:rou-annotations}
We conducted a small pilot study on RoU annotation: with the help of 5 annotators, we iteratively crafted guidelines and tested them on 12 stories. Here we share some initial qualitative observations.

For spatial annotations, agreement improved when annotators first drew a simple sketch of each scene, then translated their sketches into statements. This process seemed to help annotators notice implicit spatial facts. Some annotators also reported that sketches lowered the cognitive burden.

For temporal annotations, annotators generally agreed on what events took place and the temporal relations between them. Disagreements stemmed mainly from choices of which implicit occurrences to annotate. We are exploring ways to promote consistency, including having annotators draw timelines to draw attention to missing events. We are also looking to incorporate prior art \citep[e.g., TimeML;][]{TimeML} into our guidelines.

On causal and motivational questions, we were pleasantly surprised by the conceptual consistency between annotators. Annotators appealed to similar causal assertions, even bottoming out in similarly detailed general rules. What was less consistent was structure---how causal chains were carved into links and how bullets were nested. Annotators also occasionally omitted self-evident general rules or supporting facts. We are optimistic that both issues can be improved by more examples and training.

As expected, annotators occasionally differed on which causal contrasts to include. Such borderline judgments of salience may be inevitable, and seem to warrant use of the pyramid method.

\subsubsection{Free-text evaluation} \label{sec:free-text}
It is difficult to evaluate a system directly on an RoU or a rubric, as they are written in plain English. One option is to pose broad ToU questions (e.g., ``What events happened and in what order?'') and then to automatically compare systems' full free-text answers to annotators'. But this would require an automated comparison metric, and existing metrics such as ROUGE and BLEU are concerned only with lexical similarity. Their correlation with humans' quality judgments is substantial but not stellar \citep{BLEU-re-eval}, and high scores do not always indicate good answers in MRC \citep[see][]{RougeBleuMRC,QA-metrics}. Superficial similarity measures may prove particularly weak given how open-ended ToU questions are. 

Alternatively,  human evaluators could read both the RoU-derived rubric and the system output and decide whether the output adequately covers each nugget from the rubric. This is how the pyramid method is typically applied in summarization.

Still a third possibility is to have human evaluators ask targeted questions about each nugget from the rubric. The evaluators could then judge whether the system's shorter free-text answers 
reflect a consistent understanding of that nugget. Such evaluation would be especially powerful if the evaluators knew the NLP systems' typical shortcuts and could reword a given question accordingly: a suspicious evaluator could query for the same fact in multiple ways to verify that the system consistently gets it right. This would make results more satisfying than many MRC evaluations, as systems couldn't rely on terse answers being interpreted charitably.

Of course, using humans for the final evaluation is expensive, even if automated metrics are used during model development. Human evaluators also add variability and subjectivity, as they may probe differently for the same knowledge or find a given answer more or less convincing. Still, new tasks often start with human evaluation while the community fine-tunes what is worth measuring, and only later to progress to automated metrics that approximate human judgment. Such were the trajectories of topic model coherence \citep[see][]{AutoTopicModelEval}, summarization \citep[see][]{AutoPyramid}, and machine translation \citep[see][]{BLEU}, so it is a plausible pathway for RoU evaluation, too.

\subsubsection{Thorough multiple-choice evaluation} \label{sec:MC}
Free-response is a compelling format that is tricky to evaluate.  Multiple-choice inverts the trade-off: it is less compelling, but much easier to evaluate.

With the help of the ToU, a multiple-choice (MC) test can be fairly comprehensive. Question writers would first write out RoUs for a story, and perhaps reconcile them into a weighted rubric. They would then write MC questions targeting each nugget in the rubric: What goal is Rover pursuing by running inside rather than staying put? Where was Rover after he ran through the door? How were Rover, the house, and the rain positioned at the end of the story? Etc. Such a thorough MC test based on RoUs would be a step up from current tasks.

The downside of an MC task is that, though easy to evaluate, it would be questionable as a measure of comprehension. All MC tasks suffer from the same lack of naturalness: questions do not normally come with candidate answers, and ranking candidates is simply easier than the tasks MRC should ultimately support. Furthermore, systems learn to exploit incidental surface features in the question, sometimes performing well even without seeing the passage \citep{RCBenchmarkCritique}. When humans take MC tests, we can make strong assumptions about what they must know or do to succeed; an NLP system offers no such assurances.

In the long run, then, we do not see multiple choice as an adequate format for demonstrating MRC. 
Still, such tests offer some leverage for progress in the short term.


\subsection{Approach 2: Competing to satisfy judges}
The RoU guidelines put a stake in the ground as to how ToU questions should be answered. But as noted above, ToU questions, particularly ``why'' questions, admit many good answers. The ones canonicalized by the guidelines and by annotators following them may not always be the most useful.

Consequently, it may prove beneficial to appeal directly to human intuition about what understanding entails. We have assumed that what lets humans perform story-related tasks is that they possess some internal answers to the ToU. If we further assume that humans can be led to favor machine answers that resemble their own internal ones, then humans should make good judges of answer quality even without the guidance of RoUs. 

Accordingly, we could let humans judge system's full free-text answers based only on intuitive preferences. Evaluators could still be guided to ask ToU questions thoroughly, but extensive guidelines would not be needed: neither asking questions nor recognizing good answers demands nearly as much specification as stating canonical answers.

Whereas the approaches in \S\ref{sec:annotators} must strive for replicability in humans' answers, this approach seeks replicability only in humans' \emph{judgments} of answers. We suggest two ways to achieve this.

First, in the absence of a rubric, we suspect that answers would best be judged via pairwise comparisons. For free-text writing, humans generally find comparative assessment easier than absolute scoring \citep{ComparativeJudgment}, and comparison is already used to evaluate natural-language generation \citep[see, e.g.,][]{NlgComparison}. Comparisons also mitigate the difficulty of spotting errors of omission: when evaluators see an incomplete answer in isolation, they may gloss over or mentally fill in what was left unsaid. Comparing against a more complete competing answer makes it easier to notice gaps.

Second, evaluators can be guided to tease apart their judgments into several desirable dimensions of explanations\textemdash e.g., accuracy, depth, and coherence\textemdash just as is often done for natural language generation. Pilot studies would be required to refine the dimensions and their specifications. 

\section{Current MRC systems do not comprehend stories} \label{sec:experiments}
If current systems performed well on the ToU, our argument would be moot. This section presents evidence that they do not.

\subsection{Data and experimental setup}
To test existing systems, the questions must be presented in a form the systems can handle. Many systems were designed for span extraction, but the ToU does not lend itself to answering with text spans. Instead, we report on experiments with a pilot version of the MC task described in \S\ref{sec:MC}.

To construct the test, we selected the first two narrative stories in the dev set of \nameandcite{RACE}. Based on our preliminary annotation guidelines, one annotator read both stories, drafted an RoU for each, and wrote a question for each statement in the rough RoUs. The annotator then collaborated with several others to write distractor answers, each characterized by one or more of the following: small surface variations on the correct answer that change the meaning; language from the passage, especially words that appear near words from the question; and language that might plausibly collocate with words from the question.

As an additional test for robustness, questions came in ``variant groups'': each question was paired with a variant, or occasionally more than one, that asks for the same information in a different way (see Figure \ref{fig:race-questions}). The distractors were often altered as well. We then evaluated accuracy in two ways: counting each question independently and counting each variant group as one unit. In the latter method, the group is marked correct only if both variants were answered correctly. This simulates a suspicious evaluator re-asking the question and deducting points if the model does not consistently exhibit the desired understanding.

\begin{figure}[!tbp]
    \centering
    \small
    \begin{mdframed}
    
    \textbf{Q)} What actually happened when Mr.\ Green and the man drove together?
    \begin{enumerate}[label=\Alph*), itemsep=0pt, parsep=1pt, topsep=1pt]
        \rightanswer{\item They came to a small house.}
        \item They came to a hotel.
        \item They traveled around the country.
        \item They stopped several times at the side of the road.
    \end{enumerate}
    \vspace{0.45em}
    \textbf{Q')} How did the man's directions actually turn out?
    \begin{enumerate}[label=\Alph*), itemsep=0pt, parsep=1pt, topsep=1pt]
        \rightanswer{\item The directions the man gave led to where the man wanted to go.}
        \item The directions the man gave led to where Mr.\ Green wanted to go.
        \item The directions Mr.\ Green gave led to where the man wanted to go.
        \item The directions Mr.\ Green gave led to where Mr.\ Green wanted to go.
    \end{enumerate}

    \end{mdframed}
    \caption{An example variant group from our ToU-based questions; correct answers \rightanswer{in italics}. In the associated RACE story, a man tricks Mr.\ Green into driving him home under the pretense of guiding Mr.\ Green to a hotel. See Appendix \ref{app:mc-questions} for the full story text.}
    \label{fig:race-questions}
\end{figure}

The resulting dataset contains a total of 201 questions (98 variant groups). 29\% are spatial or temporal; the remaining 71\% are causal or motivational. The questions average 5.1 options, with a minimum of 4. (Including many distractors somewhat mitigates the weaknesses of the MC format.) All questions are included in the supplementary materials; Appendix \ref{app:mc-questions} shows many examples.

For validation, the questions were presented to two colleagues with non-technical degrees. They scored 96\% and 91\% (measured on variant groups), suggesting that motivated, well-educated humans have little trouble with our questions.

Finally, we put the questions to \nameandcite{XLNet},\footnote{For questions with more than four answers, we split the answers across multiple sub-questions, all of whose answer sets contained the correct answer. We counted the question correct only if that answer was chosen across all answer sets. Chance performance was adjusted accordingly.} a large, transformer-based language model trained with generalized autoregression on BooksCorpus and English Wikipedia. After fine-tuning, the model achieves 81.75\% on the original RACE task (within 5 points of the best non-ensemble model at the time of the experiments).

\subsection{Results and Discussion}
Our results (Table \ref{tab:results}) show that XLNet performs poorly. On individual questions, it scores just 37\%, closing less than a third of the gap between chance and human performance. This strongly suggests that whatever XLNet is doing, it is not learning the ToU's crucial elements of world understanding. 
Furthermore, the system's performance is brittle, with many correct answers attributable to luck and/or unreliable cues: when moving from questions to variant groups, human performance falls just 3 points. XLNet's performance, on the other hand, falls 17 points, which leaves the system closing just 18\% of the chance-vs.-human gap.

\begin{table}[!tbp]
    \centering
    \small
    \setlength{\tabcolsep}{9pt}
    \begin{tabular}{=l+c+c@{\hskip 5pt}+c}
        \toprule
         & \textbf{All} & \multicolumn{2}{c}{\textbf{By question type}} \\
         \cmidrule(lr{7pt}){2-2} \cmidrule(l{6pt}r{6pt}){3-4}
         & & \makecell{Spatial + \\ Temporal} & \makecell{Causal + \\ Motivational} \\
         \midrule
         Per question & 37\% & 33\% & 38\% \\
         \rowstyle{\color{gray}}
         ~~~Chance & 15\% & 20\% & 13\%  \\
         \rowstyle{\color{gray}}
         ~~~Human (avg.) & 96\% & 93\% & 97\%  \\
         Per variant group & 20\% & 14\% & 23\%  \\
         \rowstyle{\color{gray}}
         ~~~Chance & 4\% & 5\% & 5\% \\
         \rowstyle{\color{gray}}
         ~~~Human (avg.) & 93\% & 90\% & 95\%  \\
         \bottomrule
    \end{tabular}
    \caption{XLNet accuracy on our ToU-based questions.}
    \label{tab:results}
\end{table}


Although we tested only XLNet, all the other models that currently dominate the leaderboards are similar pre-trained language models; none has any distinguishing characteristic that might be expected to produce dramatically better results on our dataset. Likewise, no existing dataset is so much more systematic than RACE that fine-tuning on it should dramatically improve results on our dataset. Especially given that multiple-choice tests are artificially easy for systems (see \S\ref{sec:MC}), our pilot experiment offers strong evidence that existing MRC systems do not succeed on the ToU.

\section{Taking the ToU idea forward} \label{sec:future}
Our ToU for stories is a first attempt at defining what MRC systems should comprehend in a principled, systematic way. Drawing on work in psychology, philosophy, and pedagogy, we have argued for the ToU as a minimal standard and a valuable target for MRC. We have also shown it to be beyond the reach of current systems.

We therefore suggest that the NLP community further build on our ToU. This includes refining and perhaps expanding the questions; better defining the answers and evaluation procedures; building MRC corpora based on the ToU; and developing better-performing systems. We ourselves are working on all four, and we welcome collaboration.

But even beyond our ToU, the broader point stands: existing MRC approaches are not satisfactorily testing for a systematic set of content. Our efforts demonstrate that it is possible, with a sufficiently interdisciplinary approach, to define a plausible floor for comprehension for a given class of applications. If MRC is to achieve its ultimate goals, we\textemdash the NLP community\textemdash owe it to ourselves to ensure that our reading comprehension tests actually test for the comprehension we desire.

\bibliography{position-paper}
\bibliographystyle{acl_natbib}

\clearpage
\appendix
\section{Examples of applying the ToU to stories for applications} \label{sec:tou-examples}
In the main text (\S\ref{sec:why-stories}), we suggested that many advanced applications hinge on understanding the elements captured by our ToU for stories. Here we offer several examples from two domains.

\subsection{Law}
For the foreseeable future, legal decision-making will be the province of lawyers, not AI. However, one plausible use for MRC in a legal setting is as a \textbf{screening tool} for helping non-lawyers determine whether a case has enough merit to bother bringing in a lawyer.

For example, consider the first-person narrative below (fictional, but based on an amalgam of several real news stories):
\begin{quote}
    My property borders on public lands where hunting is allowed. Last month, a hunter tracked a buck onto my property. He claims he didn't see my boundary sign. He ended up stepping up onto the remains of an old stone wall, which crumbled, and he broke his wrist. Now he's saying I can give him \$10K now and he'll walk away, or else he's going to sue me for much more.
\end{quote}

Before contracting a lawyer, the property owner may want to assess whether there is any merit to the threat. On the other side of the deal, a law firm that offers free initial consultations may wish to avoid wasting time on cases that are clear non-starters.

A second legal application for NLU tools might be helping a lawyer \textbf{search for precedents}. For instance, a tool could help with the narrative above (or perhaps a third-person version of it) by looking for cases with similar elements---e.g., an accidental trespass resulting in injury.

To assist in such application scenarios, a system would of course need information about legal codes. But it would also have to understand what happened in the cases it is trying to analyze. To that end, the answers to ToU questions would be essential, as demonstrated in Table \ref{tab:legal-qs}. The table shows ToU questions and answers that would be key to understanding the landowner’s situation. (These questions are ones the system would answer for itself while reading, not necessarily questions it would be asked by a user.)

\begin{table*}[htb!]
    \centering
    \small
    \rowcolors{2}{gray!15}{white}
    \begin{tabular}{>{\bigstrut[t]\raggedright}p{4.6em}<{\bigstrut[b]}
                    >{\raggedright}p{8em}<{\bigstrut[b]}
                    >{\raggedright}p{16.5em}<{\bigstrut[b]}
                    >{\raggedright\arraybackslash}p{16.2em}<{\bigstrut[b]}}
        \hiderowcolors
        \toprule
        \multicolumn{1}{>{\raggedright}p{4em}}{\textbf{Question type}} & \textbf{ToU question} & \textbf{Example (partial) answer to ToU question} & \textbf{Significance to legal application} \\
        \midrule
        \multicolumn{1}{>{\raggedright}p{4em}}{Spatial} & Where was the hunter when he broke his wrist? & On the landowner's property. & The locations of events are legally relevant in many ways. For one, property owners may be held liable for injuries that occur on their property. Additionally, however, property owners may not be liable for injuries suffered by trespassers.\\
        \showrowcolors
        Spatial & Where was the boundary sign? & On the boundary between the public lands and the writer's property. & The presence of a sign may shield the landowner from responsibility, but recognizing that means understanding that it would mark the boundary between the two properties.\\
        Temporal & When did the stone wall fall into disrepair? & Sometime before the story started. & How long the wall has been in disrepair may be legally relevant. Since the exact timing was not given, the system might flag this question for further clarification.  \\
        Temporal & Has the hunter sued? & No, although he may do so in the future. & If the hunter had already sued, the landowner might need representation whether or not the suit had merit.\\
        Causal & Why did the hunter break his wrist (rather than his wrist remaining intact)? & Because he stepped onto the wall (rather than stepping elsewhere), which led to him falling (rather than remaining upright, because the wall was in disrepair rather than better condition), which led to him breaking his wrist (rather than his wrist remaining intact). & The wall's disrepair was allegedly an important causal factor in the injury, making it more plausible that the landowner could be held responsible.\\
        Motivational & Why did the hunter claim he didn't see a sign (rather than saying nothing of signs)? & He would prefer that others believe that he entered the property unwittingly (rather than deliberately), either because he in fact enter unwittingly or because he would like to deny his deliberate violation. He believes that if he says he did not see a sign, others will be more likely to believe this (whereas if he says nothing, they may assume he saw the sign). & The hunter's claim of unwitting entry could be motivated either by true innocence or by deception, which affects whether it should be believed---and unwitting entry may be treated differently by the law. The system may want to flag this claim for follow-up questions about its plausibility. \\
        Causal & Why did the hunter enter the private property (rather than stopping at the boundary)? & Possibly because the hunter didn't see the sign (rather than seeing it), so he remained unaware he was crossing the boundary (rather than realizing he was).  & There may be a mechanistic (non-motivational) explanation for why the hunter did not stop at the boundary, and again, unintentional entry may be legally different. Also, the landowner may have been responsible for posting signs that would keep people away from his property if there were any hazards.\\
        Motivational (recursive explanation for the end of the previous causal chain) & Why might being aware of the boundary have made the hunter stop, whereas being unaware of it (may have) led him to cross it? & The hunter likely prefers staying within the law to violating it. If he had known he was at the boundary of private property, he would have known that continuing past the boundary would be illegal trespass, but not knowing about the boundary meant he did not know continuing could be trespassing. & The hunter suggested that missing the sign led to accidentally entering the property, but that claim hinges on the assumption that had he known about the property line, he would have respected it. That may be a challengeable assumption.\\
        Motivational & Why did the hunter threaten to sue, rather than suing immediately? & The hunter would prefer to get less money than to possibly get more money but experience the hassle of a lawsuit and risk getting nothing. He believed that if he threatened, the property owner might be afraid of losing more money and give him the \$10,000 (whereas if the hunter sued immediately he would have no chance to avoid the hassle and risk). & It is possible that the very act of extorting money via a threat of a lawsuit has legal implications.  Also, this action by the hunter may indicate that he considers the risk of losing the case high or that he is otherwise reluctant to pursue a lawsuit, which may affect what course of action the landowner ultimately wants to take. \\\addlinespace[-2pt]
        \bottomrule
    \end{tabular}
    \caption{Example ToU questions and answers for a legal application.}
    \label{tab:legal-qs}
\end{table*}

\subsection{Medicine}
Medicine also offers ample opportunity for an MRC system competent in the narrative ToU to assist doctors and researchers. Narratives pervade electronic health records in the form of doctors' notes, which record information ranging from patient history to detailed descriptions of surgical procedures.

One narrative-based medical application is helping doctors \textbf{understand a prior doctor's rationale}. Currently, doctors often spend time sifting through a patient's records to understand why a prior doctor made a certain decision. The reasoning is often explained, but many documents must be searched to find the relevant note.

For example, consider the real medical note below,\footnote{Quoted from \url{https://www.mtsamples.com/site/pages/sample.asp?Type=96-&Sample=1939-Breast\%20Cancer\%20Followup\%20-\%201}} recorded after a routine follow-up appointment following breast cancer treatment:
\begin{quote}
    She underwent radiation treatment ending in May 2008. She then started on Arimidex, but unfortunately she did not tolerate the Arimidex and I changed her to Femara. She also did not tolerate the Femara and I changed it to tamoxifen. She did not tolerate the tamoxifen and therefore when I saw her on 11/23/09, she decided that she would take no further antiestrogen therapy. She met with me again on 02/22/10, and decided she wants to rechallenge herself with tamoxifen. When I saw her on 04/28/10, she was really doing quite well with tamoxifen. She tells me 2 weeks after that visit, she developed toxicity from the tamoxifen and therefore stopped it herself. She is not going take to any further tamoxifen.
\end{quote}
A future doctor may wonder why the patient is not on hormone therapy, which would be standard procedure. This explanatory note may be hard to find amongst the many notes in the patient's record.

\begin{table*}[htb!]
    \centering
    \small
    \rowcolors{2}{gray!15}{white}
    \newcommand{\lastcolwidth}{20.7em}
    \begin{tabular}{>{\bigstrut[t]\raggedright}p{4.6em}<{\bigstrut[b]}
                    >{\raggedright}p{5em}<{\bigstrut[b]}
                    >{\raggedright}p{15em}<{\bigstrut[b]}
                    >{\raggedright\arraybackslash}p{\lastcolwidth}<{\bigstrut[b]}}
        \hiderowcolors
        \toprule
        \multicolumn{1}{>{\raggedright}p{4em}}{\textbf{Question type}} & \textbf{ToU question} & \textbf{Example (partial) answer to ToU question} & \textbf{Significance to medical application} \\
        \midrule
        \multicolumn{1}{>{\raggedright}p{4em}}{Temporal} & When did the patient start and stop taking tamoxifen? & Multiple times: She started taking it sometime after May 2008 and stopped taking it by 11/23/09. Then, she started taking it again on 02/22/10, and stopped taking it by mid-May 2010. & \multicolumn{1}{>{\raggedright}p{\lastcolwidth}}{A clinical trial may be seeking patients who kept stopping and starting a specific drug. It may also be important how long the side effects took to develop. \par ~~~~Also note that if the question of interest is really a counting question (“how many times”), this relies most of all on an underlying temporal understanding like the one captured by the ToU.}\\
        \showrowcolors
        Causal/ Motivational & Why is the patient not taking an anti-estrogen drug (rather than taking one)? & She was taking Arimidex, and it caused strong side effects (rather than her having mild or no side effects). Preferring fewer side effects, she therefore tried Femara (rather than continuing with Arimidex). Femara also caused side effects, so for the same reasons as before, she tried switching to tamoxifen (rather than continuing the Femara), but it also caused side effects. The patient preferred not experiencing the side effects to having the medical benefits of the drugs, so she decided not to take any such drug (rather than continuing with one of the above). & \multicolumn{1}{>{\raggedright}p{\lastcolwidth}}{A future doctor may expect the patient to be on an anti-estrogen drug, as that is standard for someone with her history of breast cancer. Understanding that the patient has tried many drugs and decided to stop them may inform the doctor’s course of action. The doctor might proceed differently if he determined that she had stopped for some other reason—e.g., that she simply lapsed in a prescription. \par ~~~~Also, a clinical trial may be seeking patients who stopped taking a drug because of side effects. Furthermore, the trial might be seeking specifically patients who stopped taking the drug at the advice of the doctor.} \\\addlinespace[-2pt]
        \bottomrule
    \end{tabular}
    \caption{Example ToU questions and answers for a medical application.}
    \label{tab:medical-qs}
\end{table*}

A second medical application is \textbf{finding patients who qualify for medical trials}. For instance, a pharmaceutical company might develop a new anti-estrogen drug that they believe has milder side effects. They would then want to find patients who had already tried several anti-estrogen drugs, perhaps multiple times, and had toxicity problems with all of them. Currently, research hospitals find patients for a given clinical trial by employing humans to read through the hospital's database of medical notes and determine which patients meet the trial's criteria. 

To assist in such application scenarios, an automated system would have to understand medical notes like the one above. In the rationale-finding application, it would have to interpret the note well enough to recognize that it explains the current medical regimen; in the patient-finding application, the system would have to recognize that this patient went on and off of several anti-estrogen drugs because of side effects. Again, understanding the answers to ToU questions would be essential, as demonstrated in Table \ref{tab:medical-qs}.

\section{Example ToU-based multiple-choice questions on a RACE story} \label{app:mc-questions}
\setlist{label=\Alph*), itemsep=0pt, parsep=0pt, topsep=1pt,labelindent=7pt}
\newcommand{\question}[1]{\bigbreak\noindent\begin{minipage}{\linewidth}\noindent#1\end{minipage}\bigbreak}

\subsection{The story}
Mr.\ Green was traveling around the country in his car. One evening he was driving along a road and looking for a small hotel when he saw an old man at the side of the road. He stopped his car and said to the old man, ``I want to go to the Sun Hotel. Do you know it?''

``Yes.'' The old man answered. ``I'll show you the way.''

He got into Mr.\ Green's car and they drove for about twelve miles. When they came to a small house, the old man said, ``Stop here.''

Mr.\ Green stopped and looked at the house. ``But this isn't a hotel.'' He said to the old man.

``No,'' the old man answered, ``This is my house. And now I'll show you the way to the Sun Hotel. Turn around and go back nine miles. Then you'll see the Sun Hotel on the left.''

\subsection{RACE's original questions}

Answers marked correct by RACE are italicized.

\question{\textbf{Q1.} Where did Mr.\ Green want to sleep that night?
\begin{enumerate}
    \item In his car.
    \item In his own house.
    \rightanswer{\item In a hotel.}
    \item In the old man's house.
\end{enumerate}}

\question{\textbf{Q2.} Why did Mr.\ Green stop his car?
\begin{enumerate}
    \item Because he found a hotel.
    \item Because the lights were red.
    \rightanswer{\item Because he saw an old man.}
    \item Because he saw a friend.
\end{enumerate}}

\question{\textbf{Q3.} Where did the old man promise  to take Mr.\ Green?
\begin{enumerate}
    \item To Mr.\ Green's house.
    \item To the old man's house.
    \rightanswer{\item To the SunHotel.} [sic]
    \item To the country.
\end{enumerate}}

\question{\textbf{Q4.} Why didn't the old man stop Mr.\ Green when they passed the hotel?
\begin{enumerate}
    \item Because he wanted Mr.\ Green to sleep in his house.
    \rightanswer{\item Because he wanted to get home.}
    \item Because he didn't see the hotel.
    \item Because he didn't know the hotel.
\end{enumerate}}

\question{\textbf{Q5.} How far was it from the place where Mr.\ Green met the old man to the Sun Hotel?
\begin{enumerate}
    \item About nine miles.
    \rightanswer{\item About three miles.}
    \item About twenty-one miles.
    \item About twelve miles.
\end{enumerate}}

\subsection{A sampling of our ToU-based questions}

Correct answers are italicized. Questions are numbered with the IDs used in our dataset, which is available in this paper's supplementary data. The first number in each question ID indicates the variant group; the second number is a group-independent question index.

\subsubsection{Causal chains}

The questions below target different parts of causal chains explaining why the agents in the story took the actions that they did. The first five ask about why Mr.\ Green stopped his car (vs. continuing to drive); the next five ask about why the old man said he would show Mr.\ Green the way (vs. just giving him directions).

\question{\textbf{Q1-1.} Why did Mr.\ Green stop his car the first time?
\begin{enumerate}
    \rightanswer{\item Because if he stopped his car, he could ask the man something.}
    \item Because if he stopped his car, he could make a new friend.
    \item Because if he stopped his car, the old man could get in.
    \item Because the directions he asked for said to stop the car.
    \item Because if he stopped his car, he could drive for about twelve more miles.
    \item Because he got a flat tire.
    \item Because he was driving along a road.
    \item Because he was traveling around the country.
    \item Because he said, ``I want to go to the Sun Hotel''.
\end{enumerate}}

\question{\textbf{Q2-3.} Why did Mr.\ Green want to ask the man something?
\begin{enumerate}
    \rightanswer{\item Because there was something he didn't know.}
    \item Because he liked to ask questions.
    \item Because there was a chance to make a friend.
    \item Because he didn't want to drive past the man without helping him.
    \item Because if he stopped his car, he could drive for about twelve miles.
    \item Because he got a flat tire.
    \item Because he was driving along a road.
    \item Because he said, ``I want to go to the Sun Hotel''.
\end{enumerate}}

\question{\textbf{Q3-7.} Before they spoke at all, what did Mr.\ Green hope the man would be able to do?
\begin{enumerate}
    \rightanswer{\item Tell him where the hotel was.}
    \item Tell him where the small house was.
    \item Get in his car.
    \item Drive for about twelve miles.
    \item Take him to his house.
    \item Take him to the hotel.
    \item See an old man.
\end{enumerate}}

\question{\textbf{Q4-9.} What did Mr.\ Green hope the conversation with the old man would enable him to do?
\begin{enumerate}
    \rightanswer{\item Get where he was going}
    \item Travel around the country
    \item See what he was seeing
    \item Stop and look at a house
    \item Drive with the old man
    \item Come to a small house
    \item Turn around and go back
\end{enumerate}}

\question{\textbf{Q5-11.} What was Mr.\ Green trying to do throughout the story?
\begin{enumerate}
    \rightanswer{\item To stay at the small hotel}
    \item To drive along a road
    \item To pass the small hotel
    \item To come to a small house
    \item To see the old man
    \item To stop at the side of the road
    \item To speak with the old man
\end{enumerate}}

\question{\textbf{Q6-12.} Why did the old man make his initial offer to Mr.\ Green?
\begin{enumerate}
    \rightanswer{\item The old man was appearing to help Mr.\ Green while actually tricking him.}
    \item The old man was appearing to trick Mr.\ Green while actually helping him.
    \item Mr.\ Green was appearing to help the old man while actually tricking him.
    \item Mr.\ Green was appearing to trick the old man while actually helping him.
\end{enumerate}}

\question{\textbf{Q7-14.} Why did the old man say he would show Mr.\ Green the way instead of just giving directions?
\begin{enumerate}
    \rightanswer{\item So Mr.\ Green would let him into his car.}
    \item So Mr.\ Green would stop his car.
    \item So Mr.\ Green would say something to the old man.
    \item So he could answer Mr.\ Green.
    \item So they could go to the hotel.
    \item So Mr.\ Green would take him to the hotel.
\end{enumerate}}

\question{\textbf{Q10-20.} Where did the old man expect he and Mr.\ Green would drive together to?
\begin{enumerate}
    \rightanswer{\item The house}
    \item The Sun Hotel
    \item The side of the road
    \item Back nine miles
\end{enumerate}}

\question{\textbf{Q11-22.} Why did the man want to ride with Mr.\ Green?
\begin{enumerate}
    \rightanswer{\item He wanted to get home.}
    \item He wanted to get to the hotel.
    \item He wanted to stand at the side of the road.
    \item He wanted to answer Mr.\ Green.
    \item He wanted to get into Mr.\ Green's car.
\end{enumerate}}

\question{\textbf{Q13-26.} What is one reason the man's plan worked?
\begin{enumerate}
    \rightanswer{\item Mr.\ Green wouldn't know where they were really going.}
    \item Mr.\ Green wouldn't know what his name really was.
    \item Mr.\ Green wouldn't know how old he really was.
    \item He wanted to see the hotel on the left.
    \item He showed Mr.\ Green the way to the hotel.
\end{enumerate}}

\subsubsection{General knowledge}

For causal and motivational questions, an RoU often includes abstract general knowledge. To interrogate these components of understanding, we we wrote questions where the answer choices do not mention any of the entities in the story. Below are general knowledge questions that target the same two events as the questions immediately above.

While we thought these questions might be especially difficult, XLNet handled them about as well as the causal/motivational questions whose answer choices explicitly mentioned story entities.

\question{\textbf{Q21-44.} What is part of the reason why Mr.\ Green stopped driving when he first saw the man?
\begin{enumerate}
    \rightanswer{\item In order to ask someone a question, you have to be close to them.}
    \item In order to get where you're going, you need to stop your car.
    \item When you travel around the country, you stop your car.
    \item When the evening arrives, you drive your car home.
    \item When you're looking for a hotel, you often stop your car.
    \item People often pick up hitchhikers. 
    \item People often stop to help others.
\end{enumerate}}

\question{\textbf{Q22-47.} Why did Mr.\ Green think the man on the side of the road might be able to help him?
\begin{enumerate}
    \rightanswer{\item Often a person in a given area is familiar with the geography of that area. }
    \item Often a person in a given area gives out useful items.
    \item Often one person can give a ride to another person.
    \item Often a person on the side of the road needs help.
\end{enumerate}}

\question{\textbf{Q23-48.} Why did Mr.\ Green want to know where the hotel was?
\begin{enumerate}
    \rightanswer{\item Getting to a place usually requires knowing where the place is.}
    \item Driving around the country usually requires knowing where you are.
    \item Talking with a person usually requires seeing where they are.
    \item Getting into a car usually requires knowing where the car is.
\end{enumerate}}

\question{\textbf{Q24-51.} Why was Mr.\ Green seeking the old man's help in the first place?
\begin{enumerate}
    \rightanswer{\item People like to sleep comfortably at night.}
    \item People like to travel in a leisurely manner around the country.
    \item People like to talk amiably with each other.
    \item People like to see interesting sights on the road.
    \item People like to be driven directly to their homes.
\end{enumerate}}

\question{\textbf{Q25-52.} Why did the old man say he would show Mr.\ Green the way, the first time?
\begin{enumerate}
    \rightanswer{\item People sometimes trick others for their own gain.}
    \item People sometimes trick others in order to help them.
    \item People sometimes help others for selfless reasons.
    \item People sometimes help others for selfish reasons.
\end{enumerate}}

\question{\textbf{Q26-54.} Why did the old man first say he would show Mr.\ Green the way instead of just giving directions?
\begin{enumerate}
    \rightanswer{\item To show someone the way means going along with them whereas giving directions means just telling them information.}
    \item To show someone the way means just giving them information whereas giving directions means going along with them.
    \item Giving directions is more effective than showing someone the way.
    \item Giving directions is less effective than showing someone the way.
    \item Giving directions is more friendly than showing someone the way.
    \item Giving directions is less friendly than showing someone the way.
\end{enumerate}}

\question{\textbf{Q28-58.} Why did the old man expect to be able to control the route as he rode with Mr.\ Green?
\begin{enumerate}
    \rightanswer{\item When taking directions, people generally go where they are told to go.}
    \item When taking directions, people usually go somewhere other than where they are told to go.
    \item When on vacation, people generally follow their itineraries.
    \item When driving with strangers, people are generally very careful.
    \item When going to a small house, people generally ride together.
\end{enumerate}}

\question{\textbf{Q29-60.} What helps explain why the man wanted to accompany Mr.\ Green on his drive?
\begin{enumerate}
    \rightanswer{\item People usually want to go home at night.}
    \item People usually want to go to a hotel at night.
    \item People usually want to travel around the country.
    \item People usually want to drive with each other.
\end{enumerate}}

\question{\textbf{Q30-62.} Why did the old man trick Mr.\ Green?
\begin{enumerate}
    \rightanswer{\item Being driven home by someone is nice and convenient.}
    \item Traveling around the country with someone is fun and exciting.
    \item Stopping and looking at someone's house is interesting and enjoyable.
    \item Answering someone's questions is fulfilling and helpful.
\end{enumerate}}

\question{\textbf{Q31-64.} What is one reason the man's plan worked?
\begin{enumerate}
    \rightanswer{\item If someone is unfamiliar with an area, they won't realize if they're going the wrong way.}
    \item If someone is familiar with an area, they won't realize if they're going the wrong way.
    \item If someone is unfamiliar with an area, they will realize if they're going the wrong way.
    \item If someone is traveling around the country by car, they will drive an old man's home.
    \item If someone wants to go to a hotel, they will go to a small house first. 
\end{enumerate}}

\subsubsection{Spatio-temporal questions}

The questions below target the spatial and temporal information in the story, asking how things were physically arranged at different points in time.

\question{\textbf{Q37-76.} Who was in the car at first?
\begin{enumerate}
    \rightanswer{\item Mr.\ Green}
    \item Both Mr.\ Green and the old man
    \item The old man
    \item Neither Mr.\ Green nor the old man
\end{enumerate}}

\question{\textbf{Q38-78.} Who was in the car when Mr.\ Green drove to the small house?
\begin{enumerate}
    \rightanswer{\item Both Mr.\ Green and the old man}
    \item Mr.\ Green
    \item The old man
    \item Neither Mr.\ Green nor the old man
\end{enumerate}}

\question{\textbf{Q39-80.} Who was probably in the car when Mr.\ Green drove away from the small house?
\begin{enumerate}
    \rightanswer{\item Mr.\ Green}
    \item Both Mr.\ Green and the old man
    \item The old man
    \item Neither Mr.\ Green nor the old man
\end{enumerate}}

\question{\textbf{Q40-82.} Who was at the small house at first?
\begin{enumerate}
    \rightanswer{\item Neither Mr.\ Green nor the old man}
    \item Mr.\ Green
    \item Both Mr.\ Green and the old man
    \item The old man
\end{enumerate}}

\question{\textbf{Q41-84.} Who was at the small house when Mr.\ Green arrived there?
\begin{enumerate}
    \rightanswer{\item Both Mr.\ Green and the old man}
    \item Mr.\ Green
    \item The old man
    \item Neither Mr.\ Green nor the old man
\end{enumerate}}

\question{\textbf{Q42-86.} Who was likely at the small house a short while after the story ends?
\begin{enumerate}
    \rightanswer{\item The old man}
    \item Mr.\ Green
    \item Both Mr.\ Green and the old man
    \item Neither Mr.\ Green nor the old man
\end{enumerate}}

\question{\textbf{Q53-109.} When driving to the old man's, on which side did they pass the hotel?
\begin{enumerate}
    \rightanswer{\item The car passed the hotel on the right side of the road}
    \item The car passed the hotel on the left side of the road
    \item The car passed the house on the left side of the road
    \item The car passed the house on the right side of the road
\end{enumerate}}

\question{\textbf{Q54-111.} How were Mr.\ Green, the car, the old man, and the window probably situated when Mr.\ Green stopped to ask the man a question?
\begin{enumerate}
    \rightanswer{\item Mr.\ Green in the car, the window down, the man on the side of the road}
    \item Mr.\ Green in the car, the window down, the man in the car
    \item Mr.\ Green in the car, the window up, the man on the side of the road
    \item Mr.\ Green in the car, the window up, the man in the car
    \item Mr.\ Green out of the car, the window down, the man in the car
    \item Mr.\ Green out of the car, the window up, the man in the car
\end{enumerate}}

\question{\textbf{Q55-113.} While the two men drove to the old man's house, how was the scene likely arranged?
\begin{enumerate}
    \rightanswer{\item Mr.\ Green and the man next to each other, in the car}
    \item The man next to Mr.\ Green next to the car
    \item The car in the man and Mr.\ Green
    \item Mr.\ Green next to the man next to the car
    \item The man at his house and Mr.\ Green in the car
    \item Mr.\ Green at the hotel and the man at his house
    \item Mr.\ Green at his house and the man at the hotel
\end{enumerate}}

\question{\textbf{Q56-115.} When Mr.\ Green was actually going the right way at the end, how was the scene likely arranged?
\begin{enumerate}
    \rightanswer{\item The man at his house and Mr.\ Green in the car}
    \item Mr.\ Green and the man next to each other, in the car
    \item The man next to Mr.\ Green next to the car
    \item The car in the man and Mr.\ Green
    \item Mr.\ Green next to the man next to the car
    \item Mr.\ Green at the hotel and the man at his house
    \item Mr.\ Green at his house and the man at the hotel
\end{enumerate}}

\subsubsection{More variant groups}

As described in the paper, for each question we wrote a second version that targeted essentially the same information in a different way. Below are additional examples of such variant groups.

\question{\textbf{Q19-39.} Why could the man still help Mr. Green by showing him the way at the end of the story?
\begin{enumerate}
    \rightanswer{\item Mr.\ Green still didn't know how to get to the hotel.}
    \item Mr.\ Green still didn't know that he was at the man's house.
    \item Mr.\ Green was still looking at the house.
    \item The old man knew where Mr.\ Green's car was.
\end{enumerate}}
\question{\textbf{Q19-40.} What information was Mr.\ Green missing that the man provided when he showed him the way the second time?
\begin{enumerate}
    \rightanswer{\item Mr.\ Green didn't know how to get to the hotel.}
    \item Mr.\ Green didn't know that he was at the old man's house.
    \item Mr.\ Green didn't know who the old man was.
    \item The old man knew where Mr.\ Green's car was.
\end{enumerate}}

\question{\textbf{Q46-94.} Who was in the car just before Mr.\ Green met the old man?
\begin{enumerate}
    \rightanswer{\item Mr.\ Green}
    \item Both Mr.\ Green and the old man
    \item The old man
    \item Neither Mr.\ Green nor the old man
\end{enumerate}}

\question{\textbf{Q46-95.} Who was in the car when Mr.\ Green approached the spot where he met the old man?
\begin{enumerate}
    \rightanswer{\item Mr.\ Green}
    \item Both Mr.\ Green and the old man
    \item The old man
    \item Neither Mr.\ Green nor the old man
\end{enumerate}}

\question{\textbf{Q22-45.} Why did Mr.\ Green want to speak to the old man?
\begin{enumerate}
    \rightanswer{\item People ask questions when they lack information.}
    \item People are interested in the places they travel.
    \item People are often very curious.
    \item Old men at the side of the road sometimes know the future.
    \item People ask questions before letting people into their cars.
    \item People interrogate hitchhikers before picking them up. 
\end{enumerate}}

\question{\textbf{Q22-46.} Why did Mr.\ Green think the old man might be able to help him?
\begin{enumerate}
    \rightanswer{\item Sometimes one person has information another person doesn't.}
    \item Sometimes one person trades a car for another person's house.
    \item Sometimes one person gives a ride to another person.
    \item Sometimes one person on the side of the road gets in another person's car.
\end{enumerate}}

\end{document}